\documentclass{midl} % Include author names

\usepackage{mwe} % to get dummy images
\usepackage{float}
\usepackage{booktabs} 
\jmlrproceedings{MIDL}{Medical Imaging with Deep Learning}
\jmlrpages{}
\jmlryear{2025}

\jmlrworkshop{Short Paper -- MIDL 2025}
\jmlrvolume{-- 35}
\editors{Accepted for publication at MIDL 2025}

\title[Hierarchical Transformer]{Hierarchical Transformer for Electrocardiogram Diagnosis}

\midlauthor{\Name{Xiaoya Tang\nametag{$^{1}$}} \Email{xiaoya.tang@utah.edu}\\
\Name{Jake Berquist\nametag{$^{1,2}$}} \Email{jake.bergquist@utah.edu}\\
\Name{Benjamin A. Steinberg\nametag{$^{3}$}} \Email{benjamin.steinberg@cuanschutz.edu}\\
\Name{Tolga Tasdizen\nametag{$^{1}$}} \Email{tolga@sci.utah.edu}\\
\addr $^{1}$ Scientific Computing and Imaging Institute, University of Utah, SLC, UT, USA \\
\addr $^{2}$ Nora Eccles Harrison Cardiovascular Research and Training Institute, University of Utah, SLC, UT, USA \\
\addr $^{3}$ University of Colorado Anschutz Medical Campus, Denver, CO, USA 
} 
\begin{document}
\maketitle
\begin{abstract}
We propose a hierarchical Transformer for ECG analysis that combines depth-wise convolutions, multi-scale feature aggregation via a CLS token, and an attention-gated module to learn inter-lead relationships and enhance interpretability. The model is lightweight, flexible, and eliminates the need for complex attention or downsampling strategies. 
\end{abstract}


\section{Introduction}
 % Vision transformers(ViT)~\cite{dosovitskiy2020image} have been applied to various medical imaging analysis tasks, demonstrating remarkable potential. ViT employs uniform patch tokenization to convert inputs into sequences of equal-length feature embeddings, termed tokens. This architecture excels at capturing global dependencies within the input by combining tokenization with multi-head self-attentions(MSA). Despite their strong model capability, Transformers often lack inductive biases inherent in CNNs—such as translation equivariance, locality, and hierarchical representations—which are vital for medical data. To address this limitation, researchers explored several strategies to integrate inductive biases into Transformers, including convolutional incorporations~\cite{wu2021cvt}, local attentions~\cite{pan2023slide}, hierarchical structuring~\cite{wang2022pvt}, and auxiliary self-supervised tasks~\cite{liu2021efficient}.

Cardiovascular diseases remain a significant health threat worldwide~\cite{natarajan2020wide,hu2022transformer}. Advances in computer-aided diagnosis have aimed to enhance the accuracy of electrocardiogram (ECG) interpretation and reduce associated costs~\cite{dong2023arrhythmia}. Existed deep learning methods applied to ECG tasks often involved elaborate convolutional and recursive structures~\cite{yan2019fusing}. Given the sequential nature of ECG signals, the application of Transformers has shown promise due to their superior capacity to model long-range dependencies~\cite{natarajan2020wide}. Recent applications of Vision Transformers(ViT) in cardiac abnormality classification~\cite{natarajan2020wide}, arrhythmia detection~\cite{hu2022transformer} and phonocardiography-based valvular heart diseases detection~\cite{jamil2023efficient} have showcased their advantages in simplicity and training efficiency over CNNs and recurrent neural networks. Several recent studies have explored hierarchical transformers to address transformers' inductive bias limitations, by using shifted-window mechanisms~\cite{li2021bat},  employing transformer to bridge CNN encoders and decoders~\cite{deng2021transbridge}, integrating ResNet and channel attention~\cite{wahid2024hybrid}, and applying a deformable ViT~\cite{dong2023arrhythmia} with depthwise separable convolutions. Although depth-wise convolutions and pyramid transformers have been noted, prior works haven't fully mined the inter-lead information nor optimized the use of hierarchical features. To address these gaps, we propose a novel hierarchical Transformer for ECG classification. Our main contributions are as follows:
\begin{itemize} \item We use depth-wise convolutions to prevent the mixing of potentially important yet implicit information across leads before the transformer. This crucial aspect has been largely overlooked in previous works.
\item We employ a CLS token to aggregate task-relevant information from multi-scale representations across stages. Instead of passing all output feature embeddings to the next stage, we propagate only the CLS token. By maintaining the same CLS token across the transformer, it aggregates multi-scale information. 
\item We integrate an attention-gated module to learn inter-leads associations. This module complements our model design alongside the depth-wise encoder. \end{itemize}

\section{Methodology}
\begin{figure*}[tp]
  \setlength{\belowdisplayskip}{10pt} % Reduces space below the figure
  \centering
  \includegraphics[scale=0.72]{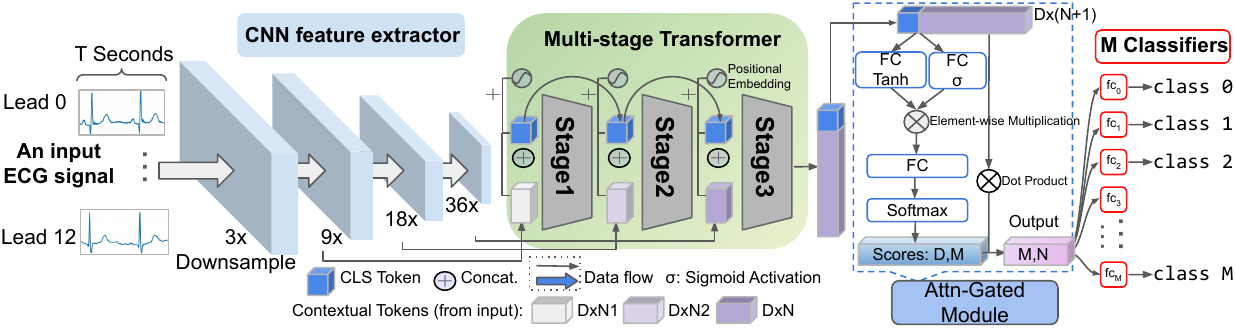}
  \caption{Framework: A Four-layer feature extractor with 1D depth-wise convolutions, three-stage transformer, and attention-gated module(from left to right). D is the embedding dimension, and N is the spatial size.}
  \vspace{-8pt} % Reduce space after the figure
  \label{fig:framework}
\end{figure*}

The framework is illustrated in Figure \ref{fig:framework}. Since the input ECG data is a multi-channel(lead) one-dimensional(1D) sequence, we employ a four-layer feature extractor consisting of 1D depth-wise convolutions~\cite{howard2017mobilenets}. Each layer uses varying kernel sizes and strides to enable progressive downsampling, adapted from~\cite{natarajan2020wide}. To incorporate multi-scale inductive biases, we design a three-stage transformer, with each stage containing a stack of MSA layers. Each stage begins by integrating a feature map(contextual tokens in Figure \ref{fig:framework}) from the CNN with a learnable CLS token and ends by passing the CLS token to the next stage, where it is integrated with another feature map of reduced spatial size. 
By adjusting the input resolutions at each stage, we manipulate the receptive field from local to global.
% We hypothesize that passing only the CLS token between stages, rather than all output embeddings, focuses the model on prediction-relevant information. Through backpropagation, the CLS token can access crucial details from the key and value representations in the attention mechanism.
Information for each lead remains distinct and uncombined throughout the transformer. To model dependencies between leads, we apply an attention-gated module comprising three linear layers (highlighted in the blue dotted rectangle in Figure~\ref{fig:framework}), inspired by~\cite{chen2022scaling}. Please refer to Appendix~\ref{sec:model} for further details. Code:\href{https://github.com/xiaoyatang/3stageFormer.git}{https://github.com/xiaoyatang/3stageFormer.git}.

\section{Results and Analysis}
% \subsection{Data and Evaluation Metrics}
We evaluate our model on two datasets: the 2020 PhysioNet/CinC Challenge dataset~\cite{alday2020classification} and the KCL potassium classification dataset from Utah. The public dataset contains 43,101 recordings, and we follow a 10-fold validation for the multi-label classification of 24 diagnoses. As in the original challenge, we report macro $F_{\beta}$, $G_{\beta}$(reflecting precision and recall), and the challenge score which penalizes incorrect diagnoses. The KCL dataset includes 54,419 training and 6,245 test recordings, with performance measured by macro-averaged AUC. Data processing and metric details are provided in Appendix~\ref{sec:data}. Our models outperform previous PhysioNet/CinC Challenge winners(Prna)~\cite{natarajan2020wide} across all metrics, rivals a two-phase method with contrastive pretraining, sCL-ST~\cite{le2023scl}, and confirm the efficacy and robustness of all variations of our model. Compared to 'Prna' which combines CNNs and transformers with a CLS token, our model's performance validates the multi-scale design and insightful use of the CLS token. We also incorporated a differential attention mechanism from language models~\cite{ye2024differential}, further enhancing performance by denoising attentions. Please refer to Appendix~\ref{sec:experiment} and ~\ref{sec:Interpretability} for detailed analysis and interpretability insights with attention maps.
\begin{table}[htbp]
  \centering
  \vspace{-10.5pt} % Reduce space above the table
  {\small
  \setlength{\tabcolsep}{3pt}  % reduce column padding
  \begin{tabular}{@{}lcccc@{}}
    \toprule
    \textbf{Model} & \textbf{Fbeta} & \textbf{Gbeta} & \textbf{Challenge} & \textbf{Size(M)} \\ 
                   % & \textbf{measure} & \textbf{measure} & \textbf{metric} & \textbf{(M)} \\ 
    \midrule
    LSTM & $0.4323\ (\pm0.55\%)$ & $0.2742\ (\pm1.90\%)$ & $0.4372\ (\pm1.67\%)$ & - \\
    CNN & $0.4519\ (\pm1.55\%)$ & $0.2862\ (\pm2.90\%)$ & $0.4542\ (\pm1.68\%)$ & - \\
    ResNet & $0.5088\ (\pm0.41\%)$ & $0.3278\ (\pm2.69\%)$ & $0.5158\ (\pm0.80\%)$ & - \\
    ViT & $0.3263\ (\pm1.65\%)$ & $0.1970\ (\pm1.88\%)$ & $0.3197\ (\pm2.44\%)$ & - \\
    Swin Transformer & $0.4812\ (\pm0.87\%)$ & $0.3045\ (\pm0.66\%)$ & $0.4811\ (\pm1.41\%)$ & - \\
    BaT~\cite{li2021bat} & $0.5011\ (\pm0.68\%)$ & $0.3125\ (\pm1.15\%)$ & $0.4958\ (\pm0.83\%)$ & - \\
    Res-SE~\cite{zhao2020adaptive} & $0.5607\ (\pm1.30\%)$ & $0.3264\ (\pm2.94\%)$ & $0.5939\ (\pm0.30\%)$ & 8.84 \\
    SpatialTemporalNet & $0.4296\ (\pm2.82\%)$ & $0.2403\ (\pm2.98\%)$ & $0.4322\ (\pm9.81\%)$ & 4.52 \\
    Prna~\cite{natarajan2020wide} & $0.4975\ (\pm5.17\%)$ & $0.2679\ (\pm6.99\%)$ & $0.5463\ (\pm3.23\%)$ & 13.64 \\
    Prna + CLS\_Token & $0.5211\ (\pm1.38\%)$ & $0.2926\ (\pm2.46\%)$ & $0.5732\ (\pm2.11\%)$ & 13.64 \\
    sCL-ST~\cite{le2023scl} & $0.5100\ (\pm-\%)$ & $\mathbf{0.3622\ (\pm-\%)}$ & $0.6053\ (\pm-\%)$ & - \\
    Ours-No Attn\_gated & $0.5672\ (\pm0.60\%)$ & $0.3296\ (\pm3.34\%)$ & $\underline{0.6174\ (\pm1.06\%)}$ & 16.62 \\
    Ours & $\underline{0.5778\ (\pm0.76\%)}$ & $0.3407\ (\pm2.18\%)$ & $0.5980\ (\pm0.85\%)$ & 16.78 \\
    Ours-No Attn\_gated-Diff & $0.5704\ (\pm1.19\%)$ & $0.3397\ (\pm0.50\%)$ & $\mathbf{0.6203\ (\pm1.23\%)}$ & 18.52 \\
    Ours-Diff & $\mathbf{0.5850\ (\pm1.15\%)}$ & $\underline{0.3459\ (\pm0.38\%)}$ & $0.6063\ (\pm0.18\%)$ & 18.63 \\
    \bottomrule
  \end{tabular}
  }
  \vspace{-2pt} % Reduce space below the table
  \caption{Performances on multi-label ECG diagnoses. Results for the first six models are reported from~\cite{li2021bat}. Others were averaged over three folds in a 10-fold cross-validation following~\cite{natarajan2020wide}, unless otherwise indicated by their sources.}
  \label{tab:1}
\end{table}

% \begin{table}[htbp]
%   \centering
%   {\small
%   \begin{tabular}{@{}lcc@{}}
%     \toprule
%     \bfseries{Model} & \bfseries{AUC} & \bfseries{Size (M)} \\
%     \midrule
%     Prna (ViT) & $0.8126 \,(\pm1.08\%)$ & 13.63 \\
%     Swin Transformer 1D-Tiny & $0.7954 \,(\pm0.11\%)$ & 47.47 \\
%     SpatialTemporalNet & $0.8218 \,(\pm0.35\%)$ & 4.41 \\
%     Res-SENet & $0.8203 \,(\pm0.09\%)$ & 8.82 \\
%     \textbf{Ours} & $\mathbf{0.8232 \,(\pm0.33\%)}$ & 10.94 \\
%     \bottomrule
%   \end{tabular}
%   }
%   \caption{Performance comparison of typical models for KCL.}
%   \label{tab:2}
% \end{table}
{
\setlength{\abovedisplayskip}{0pt}
\setlength{\belowdisplayskip}{0pt}
\setlength{\intextsep}{0pt} % Reducing space above and below the table
\begin{table}[htbp]
  \centering
  \vspace{-21pt} % Reduce space above the table
  \small
  \begin{tabular}{@{}lcc@{}}
    \toprule
    \bfseries{Model} & \bfseries{AUC} & \bfseries{Size (M)} \\
    \midrule
    Prna (ViT) & $0.8126 \,(\pm1.08\%)$ & 13.63 \\
    Swin Transformer 1D-Tiny~\cite{liu2021swin} & $0.7954 \,(\pm0.11\%)$ & 47.47 \\
    SpatialTemporalNet & $\underline{0.8218 \,(\pm0.35\%)}$ & 4.41 \\
    Res-SENet & $0.8203 \,(\pm0.09\%)$ & 8.82 \\
    \textbf{Ours} & $\mathbf{0.8232 \,(\pm0.33\%)}$ & 10.94 \\
    \bottomrule
  \end{tabular}
  \vspace{-4pt} % Reduce space below the table
  \caption{Performance comparison of typical models for KCL.}
  \label{tab:2}
\end{table}
}
\vspace{-1.5ex}
\section{Conclusion}
\vspace{-0.75ex}
Our proposed hierarchical transformer efficiently manages diverse ECG tasks, enhances interpretability by leveraging multiscale features and innovative CLS token use, and flexibly adapts to different feature extractors, attention mechanisms, and input sizes. 
\vspace{-2ex}
\midlacknowledgments{This work was supported by NIH R21HL172288. We thank Dr. Man Minh Ho for advice.}
\vspace{-0.75ex}
% \clearpage

\bibliography{midl-samplebibliography}

\begin{thebibliography}{15}
\providecommand{\natexlab}[1]{#1}
\providecommand{\url}[1]{\texttt{#1}}
\expandafter\ifx\csname urlstyle\endcsname\relax
  \providecommand{\doi}[1]{doi: #1}\else
  \providecommand{\doi}{doi: \begingroup \urlstyle{rm}\Url}\fi

\bibitem[Alday et~al.(2020)Alday, Gu, Shah, Robichaux, Wong, Liu, Liu, Rad, Elola, Seyedi, et~al.]{alday2020classification}
Erick A~Perez Alday, Annie Gu, Amit~J Shah, Chad Robichaux, An-Kwok~Ian Wong, Chengyu Liu, Feifei Liu, Ali~Bahrami Rad, Andoni Elola, Salman Seyedi, et~al.
\newblock Classification of 12-lead ecgs: the physionet/computing in cardiology challenge 2020.
\newblock \emph{Physiological measurement}, 41\penalty0 (12):\penalty0 124003, 2020.

\bibitem[Chen et~al.(2022)Chen, Chen, Li, Chen, Trister, Krishnan, and Mahmood]{chen2022scaling}
Richard~J Chen, Chengkuan Chen, Yicong Li, Tiffany~Y Chen, Andrew~D Trister, Rahul~G Krishnan, and Faisal Mahmood.
\newblock Scaling vision transformers to gigapixel images via hierarchical self-supervised learning.
\newblock In \emph{Proceedings of the IEEE/CVF Conference on Computer Vision and Pattern Recognition}, pages 16144--16155, 2022.

\bibitem[Deng et~al.(2021)Deng, Meng, Gao, Bridge, Shen, Lip, Zhao, and Zheng]{deng2021transbridge}
Kaizhong Deng, Yanda Meng, Dongxu Gao, Joshua Bridge, Yaochun Shen, Gregory Lip, Yitian Zhao, and Yalin Zheng.
\newblock Transbridge: A lightweight transformer for left ventricle segmentation in echocardiography.
\newblock In \emph{Simplifying Medical Ultrasound: Second International Workshop, ASMUS 2021, Held in Conjunction with MICCAI 2021, Strasbourg, France, September 27, 2021, Proceedings 2}, pages 63--72. Springer, 2021.

\bibitem[Dong et~al.(2023)Dong, Zhang, Qiu, Wang, and Yu]{dong2023arrhythmia}
Yanfang Dong, Miao Zhang, Lishen Qiu, Lirong Wang, and Yong Yu.
\newblock An arrhythmia classification model based on vision transformer with deformable attention.
\newblock \emph{Micromachines}, 14\penalty0 (6):\penalty0 1155, 2023.

\bibitem[Howard(2017)]{howard2017mobilenets}
Andrew~G Howard.
\newblock Mobilenets: Efficient convolutional neural networks for mobile vision applications.
\newblock \emph{arXiv preprint arXiv:1704.04861}, 2017.

\bibitem[Hu et~al.(2022)Hu, Chen, and Zhou]{hu2022transformer}
Rui Hu, Jie Chen, and Li~Zhou.
\newblock A transformer-based deep neural network for arrhythmia detection using continuous ecg signals.
\newblock \emph{Computers in Biology and Medicine}, 144:\penalty0 105325, 2022.

\bibitem[Jamil and Roy(2023)]{jamil2023efficient}
Sonain Jamil and Arunabha~M Roy.
\newblock An efficient and robust phonocardiography (pcg)-based valvular heart diseases (vhd) detection framework using vision transformer (vit).
\newblock \emph{Computers in Biology and Medicine}, 158:\penalty0 106734, 2023.

\bibitem[Le et~al.(2023)Le, Truong, Brijesh, Adjeroh, and Le]{le2023scl}
Duc Le, Sang Truong, Patel Brijesh, Donald~A Adjeroh, and Ngan Le.
\newblock scl-st: Supervised contrastive learning with semantic transformations for multiple lead ecg arrhythmia classification.
\newblock \emph{IEEE journal of biomedical and health informatics}, 27\penalty0 (6):\penalty0 2818--2828, 2023.

\bibitem[Li et~al.(2021)Li, Li, Wei, Sun, Wei, Li, and Qian]{li2021bat}
Xiaoyu Li, Chen Li, Yuhua Wei, Yuyao Sun, Jishang Wei, Xiang Li, and Buyue Qian.
\newblock Bat: Beat-aligned transformer for electrocardiogram classification.
\newblock In \emph{2021 IEEE International Conference on Data Mining (ICDM)}, pages 320--329. IEEE, 2021.

\bibitem[Liu et~al.(2021)Liu, Lin, Cao, Hu, Wei, Zhang, Lin, and Guo]{liu2021swin}
Ze~Liu, Yutong Lin, Yue Cao, Han Hu, Yixuan Wei, Zheng Zhang, Stephen Lin, and Baining Guo.
\newblock Swin transformer: Hierarchical vision transformer using shifted windows.
\newblock In \emph{Proceedings of the IEEE/CVF international conference on computer vision}, pages 10012--10022, 2021.

\bibitem[Natarajan et~al.(2020)Natarajan, Chang, Mariani, Rahman, Boverman, Vij, and Rubin]{natarajan2020wide}
Annamalai Natarajan, Yale Chang, Sara Mariani, Asif Rahman, Gregory Boverman, Shruti Vij, and Jonathan Rubin.
\newblock A wide and deep transformer neural network for 12-lead ecg classification.
\newblock In \emph{2020 Computing in Cardiology}, pages 1--4. IEEE, 2020.

\bibitem[Wahid et~al.(2024)Wahid, Mingliang, Ayoub, Husssain, Li, and Shi]{wahid2024hybrid}
Junaid~Abdul Wahid, Xu~Mingliang, Muhammad Ayoub, Shabir Husssain, Lifeng Li, and Lei Shi.
\newblock A hybrid resnet-vit approach to bridge the global and local features for myocardial infarction detection.
\newblock \emph{Scientific Reports}, 14\penalty0 (1):\penalty0 4359, 2024.

\bibitem[Yan et~al.(2019)Yan, Liang, Zhang, and Liu]{yan2019fusing}
Genshen Yan, Shen Liang, Yanchun Zhang, and Fan Liu.
\newblock Fusing transformer model with temporal features for ecg heartbeat classification.
\newblock In \emph{2019 IEEE International Conference on Bioinformatics and Biomedicine (BIBM)}, pages 898--905. IEEE, 2019.

\bibitem[Ye et~al.(2024)Ye, Dong, Xia, Sun, Zhu, Huang, and Wei]{ye2024differential}
Tianzhu Ye, Li~Dong, Yuqing Xia, Yutao Sun, Yi~Zhu, Gao Huang, and Furu Wei.
\newblock Differential transformer.
\newblock \emph{arXiv preprint arXiv:2410.05258}, 2024.

\bibitem[Zhao et~al.(2020)Zhao, Fang, Relton, Yan, Liu, Li, Qin, and Wong]{zhao2020adaptive}
Zhibin Zhao, Hui Fang, Samuel~D Relton, Ruqiang Yan, Yuhong Liu, Zhijing Li, Jing Qin, and David~C Wong.
\newblock Adaptive lead weighted resnet trained with different duration signals for classifying 12-lead ecgs.
\newblock In \emph{2020 Computing in Cardiology}, pages 1--4. IEEE, 2020.

\end{thebibliography}
% \clearpage
\appendix
\section{Model Architecture}
\label{sec:model}
\subsection{Depthwise Convolutional Feature Encoder} 
Depthwise convolutions~\cite{howard2017mobilenets} employ a distinct filter for each input channel, capturing spatial relationships without cross-channel interactions. In the context of multi-lead ECG signals, these convolutions are applied individually to each lead. Subsequently, the resulting feature maps from each lead can be transformed separately onto a new space~\cite{dong2023arrhythmia}. 
\subsection{Three-Stage Transformer} 
According to previous researches, the effective receptive field of ViT shifts from local to global as it progresses through the layers. We structure a transformer encoder into three stages, each containing a stack of MSA layers, with the division of layers tailored to specific needs. Our approach involves feeding hierarchical feature embeddings(called contextual tokens in \ref{fig:framework}) into three stages, derived from different layers of our convolutional encoder using three distinct downsampling rates from the input ECG segment. After each stage, the CLS token is extracted and concatenated with a new sequence of embeddings at a larger downsampling rate, then passed into the next transformer stage. This progressive feeding of downsampled features compels the model to transition its focus from detailed to more abstract, global patterns. Utilizing the CLS token allows us to efficiently aggregate and transfer multi-scale information to the final classification layer.

\subsection{Attention-Gated Module}
Given an output from three-stage transformer \( x \) with dimensions \( x \in \mathbb{R}^{B \times C \times S} \), where \( B \) represents the batch size, \( C \) the number of channels, and \( S \) the sequence length. The information for each lead remains distinct and uncombined. Thus we utilize an attention-gated module to model dependencies between leads, inspired by ~\cite{chen2022scaling}. This module comprises three linear layers designed to uncover latent dependencies between channels, which correspond to associations between ECG leads in this context. The attention score \( a \) is computed through an element-wise multiplication of the query and key vectors, resulting in \( a \in \mathbb{R}^{B \times C \times S} \), as shown in Eq.~\ref{Eq:1}. \( W_q \in \mathbb{R}^{S \times S} \), \( b_q \in \mathbb{R}^S \),\( W_k \in \mathbb{R}^{S \times S} \), and \( b_k \in \mathbb{R}^S \) represent the weights and biases of the linear layer for learning query and key, with \( \sigma \) denotes the Sigmoid function.
\begin{equation}
\begin{split}
    q &= \tanh(W_q x + b_q)\\
    k &= \sigma(W_k x + b_k)\\
    a &= q \odot k
\end{split}
    \label{Eq:1}
\end{equation}
A linear project is applied to the attention scores resulting in \( a' \in \mathbb{R}^{B \times C \times N} \), where \( N \) is the number of classes. The raw attentions are then normalized by a softmax and multiplied with the output from the three-stage transformer,yielding \( v \in \mathbb{R}^{B \times N \times S} \), shown in Eq.\ref{Eq:2}. These operations are analogous to the MSA mechanism. Finally, a separate classifier for each class is applied across the sequences, where \( v_i \) denotes the segment of \( v \) corresponding to the \( i \)-th class.
\begin{equation}
    \begin{split}
    a' &= \text{Projection}(a)\\
    a'' &= \text{softmax}(\text{transpose}(a', (0, 2, 1)))\\
    v &= a'' @ x\\
    \text{logits}_i &= W_i v_{i} + b_i \quad \text{for each } i \in \{1, \dots, N\}
    \end{split}
    \label{Eq:2}
\end{equation}

\section{Data and Evaluation Metrics}
\label{sec:data}
We utilize the public training data from the 2020 PhysioNet/CinC Challenge~\cite{alday2020classification} and KCL data from our group. The public dataset comprises $43,101$ recordings, and we adopt the 10-fold split used by the winner model 'Prna'~\cite{natarajan2020wide}. This setup involves a multi-label classification task related to 24 diagnoses. Following the preprocessing steps of 'Prna', we resample all recordings to $500Hz$, apply an FIR bandpass filter, and perform normalization. We also randomly crop multiple fixed-length ECG segments of $T=15$ seconds from the input, adding padding when necessary for segments shorter than $15$s.  We also leveraged the wide features that they used. For evaluation metrics, we report macro $F_{\beta}$, $G_{\beta}$, geometric mean(GM) combining precision and recall and the challenge score defined by the challenge organizers~\cite{alday2020classification}, detailed in Eq.~\ref{Eq:3}. The score $S$ generalizes standard accuracy by fully crediting correct diagnoses and penalizing incorrect ones based on the similarity between arrhythmia types. Here $a_{ij}$ represents an entry in the confusion matrix corresponding to the number of recordings classified as class $c_i$ but actually belonging to class $c_j$, with different weights $w_{ij}$ assigned based on the similarity of classes $c_i,c_j$:
\begin{equation}
    \begin{split}
        F_{\beta} &= (1+\beta^2)\cdot \frac{TP }{(1+\beta^2) \cdot TP+ FP+ \beta^2 FN}\\
        G_{\beta} &=  \frac{TP }{ TP+ FP+ \beta FN}, \beta = 2\\
        % GM &= \sqrt{F_{\beta} \cdot G_{\beta} }, \beta = 2\\
        S &= \sum_{ij} w_{ij}a_{ij}
    \end{split}
    \label{Eq:3}
\end{equation}
For the KCL potassium classification, all recordings maintain a uniform sampling rate of $500Hz$. After applying normalization, we randomly crop these to fixed segments of $T=5s$. The dataset includes $54,419$ recordings for training and $6,245$ for testing. We report the macro-averaged area under the receiver operating characteristic curve (AUC) on test data.
\section{Result Analysis}
\label{sec:experiment}
Our model surpasses previous winners in the 2020 PhysioNet/CinC Challenge dataset, Prna~\cite{natarajan2020wide} and Res-SENet~\cite{zhao2020adaptive}, and other commonly used architectures. Results are shown in~\ref{tab:1}. An important observation from these results is the enhanced performance of the standard ViT model, Prna, upon integration of a CLS token, which validates the CLS token’s significance in classification tasks. By adjusting the input resolutions at each transformer stage, we manipulate the receptive field of attention—from local, with larger spatial-sized embeddings, to global, with smaller spatial-sized embeddings. These variations in granularity are achieved by modifying the kernel sizes and strides in the CNN feature extractor. We hypothesize that passing only the CLS token between stages, rather than all output embeddings, focuses the model on prediction-relevant information. Through backpropagation, the CLS token can access crucial details from the key and value representations in the attention mechanism. By maintaining the same CLS token across the transformer, it aggregates multi-scale information, thus enhancing the model's inductive bias. By outperforming the integration of Prna with a CLS token, we confirm our model's effectiveness. Additionally, our model demonstrates that both with and without the Attention\_gated module, it maintains competitive performance, demonstrating the efficiency of our three-stage transformer.

To further validate the efficiency and generalizability of our approach, we conducted additional tests on the KCL binary classification task, comparing our model against other prominent architectures. Our model showcased the highest AUC, outperforming models such as SpatialTemporalNet and ViT (Prna), shown in~\ref{tab:2}. These results confirm the robustness and adaptability of our model, effectively identifying complex patterns essential for precise ECG classification. 
\subsection{Differential Attention Mechanism}
Inspired by attention denoising techniques in large language models and the similar sequential nature of ECG signals and language data, we adapted the differential attention mechanism from \cite{ye2024differential}. This mechanism enhances focus on relevant contexts while suppressing noise by subtracting two softmax-transformed attention maps, which reduces noise and promotes the development of distinct attention patterns, as delineated in Eq.\ref{Eq:4}. Given an input $X \in \mathbb{R}^{N \times d}$, we project it to query, key, and value tensors $\mathbf{Q}_1, \mathbf{Q}_2, \mathbf{K}_1, \mathbf{K}_2, \mathbf{V} \in \mathbb{R}^{N \times 2D}$. The differential attention operation $\text{DiffAttn}(\cdot)$ then computes the outputs. A significant adaptation for ECG analysis involves omitting the upper diagonal mask, originally used in language generative models to prevent attending to future tokens, thus allowing the model to consider the entire ECG sequence simultaneously.
\begin{equation}
\begin{split}
[\mathbf{Q}_1; \mathbf{Q}_2] &= XW^Q, \quad [\mathbf{K}_1; \mathbf{K}_2] = XW^K, \quad \mathbf{V} = XW^V \\
\text{DiffAttn}(X) &= \left(\text{softmax}\left(\frac{\mathbf{Q}_1\mathbf{K}_1^T}{\sqrt{D}}\right) - \lambda \text{softmax}\left(\frac{\mathbf{Q}_2\mathbf{K}_2^T}{\sqrt{D}}\right)\right)\mathbf{V}
\end{split}
    \label{Eq:4}
\end{equation}

\section{Interpretability}
\label{sec:Interpretability}
\begin{figure}[H]  % "Here" - Force the figure to be placed exactly here
\begin{minipage}[b]{1.0\linewidth}
  \centering
  \caption{Attentions in an abnormal case (high potassium) for leads 1 to 4, illustrating final stage attentions.}
  \label{fig:attentions_leads1-4}
  \includegraphics[height =0.4\textwidth, width=0.85\textwidth]{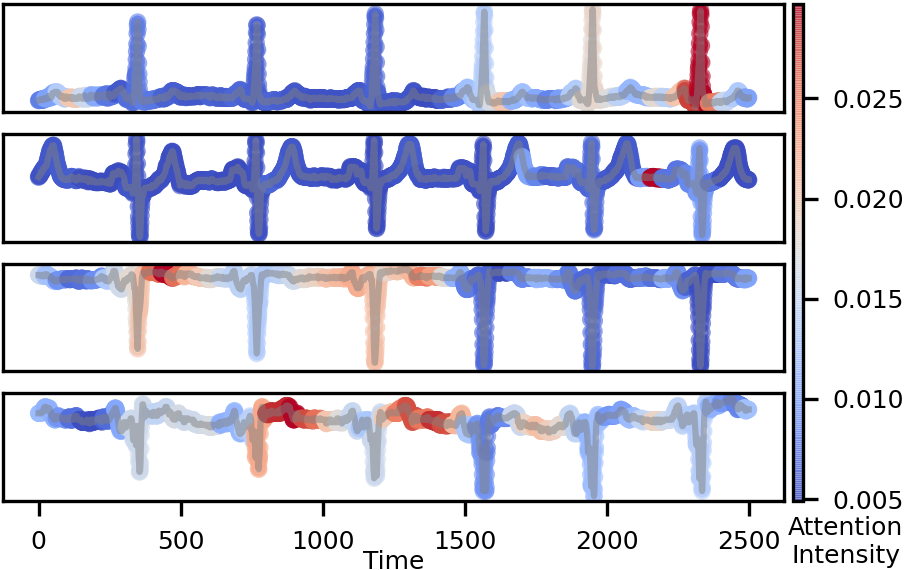}
  % best 0.55/0/97
\end{minipage}
\end{figure}

\begin{figure}[H]  % "Here" - Force the figure to be placed exactly here
\begin{minipage}[b]{1.0\linewidth}
  \centering
  \caption{Attentions in an abnormal case (high potassium) for leads 5 to 8, illustrating final stage attentions.}
  \label{fig:attentions_leads5-8}  % Ensure label is after caption
  \includegraphics[height =0.4\textwidth, width=0.85\textwidth]{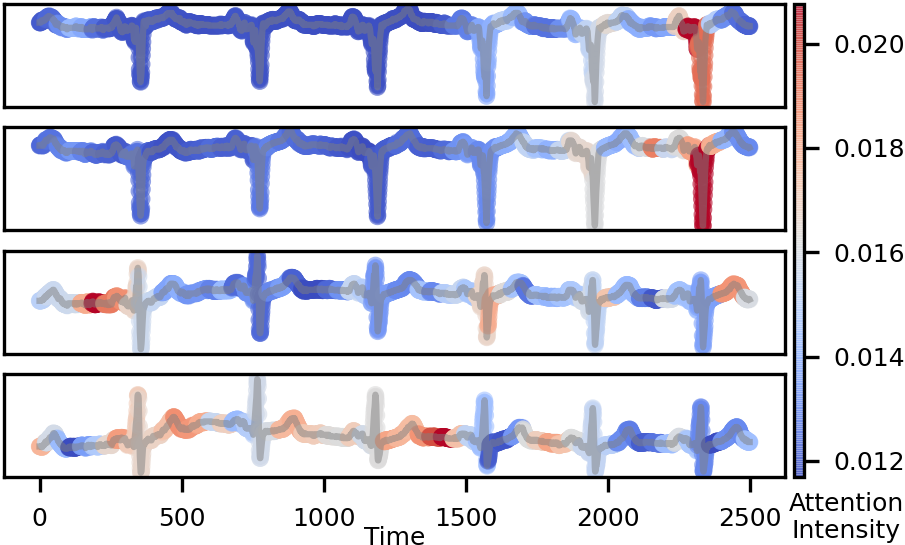}
\end{minipage}
\end{figure}

The multi-head self-attention(MSA) allows each head to learn distinct attention patterns across the time sequence. These patterns can be analogized to distinct attentions across different ECG leads, facilitated by our depthwise encoder. During the evaluation phase of the KCL classification, we randomly selected an abnormal sample, with the attention map at the final stage shown in~\ref{fig:attentions_leads1-4} and~\ref{fig:attentions_leads5-8}. We qualitatively assessed the attention map by examining which areas of the ECG signals garnered the highest attention in this unhealthy case. Notably, our model exhibited heightened attention to clinically significant features such as the QRS complex, S-T segment, and T-wave, which are recognized as clinical indicators of changes in serum potassium levels. The proposed approach also underscores how the model's attention shifts across different stages. While we do not leverage lead attention here, the attn\_gated module after MSA allowed us to discern dependencies among multiple leads. This capability further provides valuable insights into how the model relies on different leads, enhancing our understanding of deep learning models for ECG diagnosis. 


\end{document}